\documentclass[3p,times,procedia]{elsarticle}
\usepackage{ecrc}
\usepackage[bookmarks=false]{hyperref}
    \hypersetup{colorlinks,
      linkcolor=blue,
      citecolor=blue,
      urlcolor=blue}
\usepackage{amsmath}
\usepackage{listings}
\usepackage{xcolor}
\usepackage{booktabs}

\lstdefinestyle{codestyle}{
  basicstyle=\ttfamily\small,
  breaklines=true,
  frame=single,
  xleftmargin=2em,
  framexleftmargin=1.5em,
  numbers=none,
  tabsize=2,
  showstringspaces=false,
  keywordstyle=\color{blue},
  commentstyle=\color{gray},
  stringstyle=\color{red!70!black},
}

\volume{00}

\firstpage{1}

\journalname{Procedia Computer Science}

\runauth{A.V. Kozachok et al.}

\jid{procs}

\usepackage{amssymb}

\usepackage[figuresright]{rotating}

\begin{document}
\begin{frontmatter}

\dochead{30th International Conference on Knowledge-Based and Intelligent Information \& Engineering Systems (KES 2026)}%

\title{Text2DSL: LLM-Based Code Generation for Domain-Specific Languages}

\author[a]{Kozachok Alexander\corref{cor1}}
\author[b]{Nazimov Alexander}
\author[a]{Magomedov Shamil}

\address[a]{RTU MIREA --- Russian Technological University, 78 Vernadsky Avenue, Moscow 119454, Russia}
\address[b]{Academy of the Federal Guard Service of the Russian Federation, 35 Priborostroitelnaya Street, Oryol 302020, Russia}

\begin{abstract}
Domain-specific languages (DSLs) are widely used for managing operating system security policies, yet manually authoring rules in such languages demands high expertise and is error-prone. This paper formalises the task of automatic DSL code generation from natural language descriptions --- \emph{Text2DSL} --- as a distinct problem class, separate from \emph{Text-to-SQL} and general-purpose code generation. We introduce the \textbf{PolkitBench} dataset comprising 4,204 verified natural-language--to--Polkit-rule pairs, each validated through a three-level AST-based pipeline. Controlled prompt experiments on two MoE models of different scale and provenance --- GigaChat-10B-A1.8B (1.8B active parameters) and Nemotron-3-Nano-30B-A3B (3B active) --- demonstrate the critical role of structured context (BNF grammar, API specification, permitted identifier vocabulary) for LLM-based DSL code generation. Across both models, supplying context raises syntactic validity to 98.6--99.4\%, structural validity by +9.7 to +35.5~pp, and the CodeBLEU score by +60\% to +95\%. The consistency of the effect across models of different scale and provenance indicates that, for the Text2DSL class of problems, injecting a formal target-language specification into the prompt context is a robust enabling factor for high-quality generation without model fine-tuning.
\end{abstract}

\cortext[cor1]{Corresponding author. Tel.: +7-920-284-5743.}
\end{frontmatter}

\email{kozachok\_a@mirea.ru}

\section{Introduction}
\label{sec:intro}

Domain-specific languages (DSLs) play a central role in the configuration and enforcement of security policies across modern computing infrastructure. In operating systems, Polkit governs privilege escalation for desktop and server operations in Linux~\cite{freedesktop2024}, SELinux enforces mandatory access control through type-enforcement rules containing thousands of type identifiers~\cite{Smalley2001}, and AppArmor confines applications via path-based profiles. In the cloud-native domain, OPA Rego~\cite{OPA2024} defines admission control and authorisation policies for Kubernetes and microservice meshes, while Terraform HCL~\cite{Terraform2024} and AWS CloudFormation describe infrastructure-as-code. The common property of these languages is that they occupy a narrow syntactic niche --- each is described by a compact grammar --- yet demand expert knowledge of both the DSL syntax and the underlying system semantics. A single misconfigured Polkit rule can grant unintended root access; an incorrect SELinux policy can silently deny legitimate operations or, conversely, open a privilege escalation vector.

The cost of manual DSL rule authoring is well documented. Studies on security misconfiguration~\cite{Rahman2019} show that infrastructure-as-code scripts contain on average 13--31 security smells per 1,000 lines, and that policy languages are among the most error-prone artefacts in DevOps pipelines. Despite this, tool support for automatic DSL code generation lags significantly behind general-purpose programming languages.

Recent advances of large language models (LLMs) in general-purpose code generation --- demonstrated by Codex on the HumanEval benchmark~\cite{Chen2021} (pass@1 = 28.8\%), StarCoder~\cite{Li2023}, and CodeBERT~\cite{Feng2020} --- and in the Text-to-SQL task~\cite{Yu2018,Pourreza2023} suggest that DSL rule generation from natural language descriptions can be automated as well. However, there is a fundamental difference between general-purpose code generation and DSL code generation: a target DSL typically possesses a fixed and compact formal grammar, a bounded and closed vocabulary of permitted identifiers~$V$, and deterministic semantics unambiguously defined by the specification. These properties allow the DSL code generation task to be formulated as a mapping from a natural language query to a program satisfying strict formal constraints --- which we denote by the term \emph{Text2DSL}.

Despite the practical importance of DSL code generation, existing work has not explicitly formalised it as a distinct problem class with dedicated evaluation metrics and benchmarks. While prior studies addressed specific DSLs --- Pandas API~\cite{Jain2022}, Bash~\cite{Narayan2022}, Ansible~\cite{Pujar2023} --- they treated DSL generation as an application of general code generation rather than as a structurally distinct task. Existing benchmarks target either general-purpose languages (HumanEval~\cite{Chen2021}, MBPP~\cite{Austin2021}) or SQL (Spider~\cite{Yu2018}, Seq2SQL/WikiSQL~\cite{Zhong2017}), leaving DSLs with closed grammars and fixed vocabularies without a unified evaluation framework.

The contributions of this paper are as follows:
\begin{enumerate}
\item \textbf{Formalisation of the Text2DSL task} as a distinct class of code generation problems with formal requirements for syntactic and semantic correctness.
\item \textbf{Introduction of the PolkitBench dataset} --- 4,204 natural-language--to--Polkit-rule pairs verified through a three-level AST-based validation pipeline.
\item \textbf{Empirical evidence on two MoE models of different scale and provenance} (GigaChat-10B-A1.8B and Nemotron-3-Nano-30B-A3B) that structured context comprising a BNF grammar, an API specification, and a permitted identifier vocabulary is a critical factor for the quality of DSL code generation without model fine-tuning.
\end{enumerate}

\section{Related Work}
\label{sec:related}

\subsection{LLM-Based Code Generation}
\label{sec:rw_codegen}

The application of LLMs to code generation has been evaluated primarily on general-purpose language benchmarks. Codex~\cite{Chen2021} achieved pass@1~=~28.8\% on HumanEval (164 Python problems); subsequent models --- StarCoder~\cite{Li2023}, Code Llama~\cite{Roziere2023}, and DeepSeek-Coder~\cite{Guo2024} --- raised pass@1 above 50\%. The MBPP benchmark~\cite{Austin2021} extended evaluation to 974 crowd-sourced Python tasks. CodeBERT~\cite{Feng2020} and GraphCodeBERT~\cite{Guo2021} introduced pre-training objectives that capture code structure. However, all these benchmarks target languages with open grammars and unbounded vocabularies, making them unsuitable for evaluating DSL code generation, where the correctness criteria are fundamentally different (Section~\ref{sec:distinction}).

\subsection{Text-to-SQL and Schema Linking}
\label{sec:rw_sql}

The Text-to-SQL task is the closest analogue to Text2DSL. The Spider benchmark~\cite{Yu2018} (10,181 queries, 200 databases) established the standard evaluation setting; WikiSQL~\cite{Zhong2017} provided a larger but simpler single-table corpus. DIN-SQL~\cite{Pourreza2023} demonstrated that decomposing the task and injecting the database schema into the prompt (schema linking) substantially improves accuracy. PICARD~\cite{Scholak2021} introduced incremental constrained decoding that prunes syntactically invalid tokens during generation. RESDSQL~\cite{Li2023sql} separated schema linking from SQL generation, achieving state-of-the-art results on Spider. The Text2DSL task shares the closed-vocabulary property with Text-to-SQL (table/column names $\leftrightarrow$ \texttt{action.id} identifiers), but differs in grammar compactness ($\sim$10 vs.\ $\sim$50 productions) and the absence of a data-access layer for verification.

\subsection{DSL and Configuration Code Generation}
\label{sec:rw_dsl}

Work on DSL-specific code generation remains sparse. Jigsaw~\cite{Jain2022} applied LLMs to generate code using the Python Pandas API, employing post-processing and test-based filtering. Narayan et al.~\cite{Narayan2022} studied LLM-based generation of Bash commands from natural language, reporting that API documentation in the prompt reduces hallucination rates. For infrastructure-as-code, Pujar et al.~\cite{Pujar2023} fine-tuned models on Ansible playbooks. Rahman et al.~\cite{Rahman2019} analysed security smells in IaC scripts, highlighting the need for automated policy generation. While the above works address DSL-adjacent tasks, they treat code generation as an application of existing NL$\to$Code methods rather than formalising DSL generation as a structurally distinct problem class with dedicated evaluation metrics. We are not aware of a verified benchmark for a security-policy DSL in the published literature.

\subsection{Grammar-Constrained Decoding}
\label{sec:rw_grammar}

Synchromesh~\cite{Poesia2022} proposed constraining the LLM decoder to emit only tokens consistent with a target grammar, achieving near-perfect syntactic validity. GCD (Grammar-Constrained Decoding)~\cite{Geng2024} generalised this to arbitrary context-free grammars. These approaches are complementary to context injection: grammar-constrained decoding ensures syntactic correctness at the token level, while context injection provides semantic grounding (correct identifiers, API usage). Our work focuses on the latter mechanism.

\section{Problem Statement: Text2DSL}
\label{sec:problem}

\subsection{Formal Definition}
\label{sec:formal}

Let $\mathcal{Q}$ denote the set of natural language queries, $G$ the formal grammar of the target DSL, and $V$ the closed vocabulary of permitted identifiers (API names, constants, enumerations). The \emph{Text2DSL} task is defined as the construction of a mapping $f\colon \mathcal{Q} \to \mathcal{P}_G$, where $\mathcal{P}_G$ is the set of programs admitted by grammar~$G$, such that for an arbitrary query $q \in \mathcal{Q}$ the generated program $p = f(q)$ satisfies two conditions:

\begin{enumerate}
\item \textbf{Syntactic correctness:} $\mathrm{parse}(p, G) = \mathrm{OK}$, i.e., the program~$p$ admits parsing into an abstract syntax tree (AST) according to grammar~$G$.
\item \textbf{Semantic correctness:} $\mathrm{ids}(p) \subseteq V$, i.e., all identifiers used in~$p$ belong to the vocabulary~$V$.
\end{enumerate}

\subsection{Distinction Between DSL and General-Purpose Code}
\label{sec:distinction}

The Text2DSL task differs fundamentally from the general code generation task (NL$\to$Code) in three properties of the target language:

\textbf{(a) Fixed grammar.} The target DSL is described by a compact BNF grammar containing a finite number of productions. For Polkit, the grammar comprises 10~productions (Table~\ref{tab:bnf}), whereas the Python grammar expressed in PEG notation contains over 300~rules.

\textbf{(b) Closed identifier vocabulary.} The set~$V$ is fixed and known a~priori. For Polkit, $|V| \approx 65$ (categories: Systemd, login1, PackageKit, udisks2, NetworkManager, Bluetooth, CUPS, etc.). Any identifier $v \not\in V$ constitutes an error, whereas for Python $|V|$ is potentially unbounded.

\textbf{(c) Deterministic semantics.} Each API element has a single interpretation defined by the specification. For example, \texttt{polkit.Result.AUTH\_SELF} means exclusively ``prompt for the current user's password'' and cannot be used in any other context.

These properties make the Text2DSL task simultaneously \emph{simpler} in terms of the dimensionality of the admissible solution space and \emph{stricter} in terms of correctness requirements: unlike general-purpose code, DSL code admits complete formal verification of the output.

\begin{table}[h]
\caption{Fragment of the Polkit DSL BNF grammar.}
\label{tab:bnf}
\begin{tabular*}{\hsize}{@{\extracolsep{\fill}}ll@{}}
\toprule
Production & Rule \\
\colrule
\texttt{<rule>} & \texttt{polkit.addRule(function(action, subject) \{ <body> \});} \\
\texttt{<body>} & \texttt{<statement> | <statement> <body>} \\
\texttt{<statement>} & \texttt{<if-block> | <return>} \\
\texttt{<if-block>} & \texttt{if ( <condition> ) \{ <body> \}} \\
\texttt{<condition>} & \texttt{<action-check> | <subject-check> | <condition> \&\& <condition>} \\
\texttt{<action-check>} & \texttt{action.id == "<action-id>"} \\
\texttt{<subject-check>} & \texttt{subject.user == "<string>" | subject.local | subject.active} \\
 & \texttt{| subject.isInGroup("<string>")} \\
\texttt{<return>} & \texttt{return polkit.Result.<result>;} \\
\texttt{<result>} & \texttt{YES | NO | AUTH\_SELF | AUTH\_ADMIN} \\
\botrule
\end{tabular*}
\end{table}

\subsection{Polkit as a Representative Instance of the DSL Class}
\label{sec:polkit}

Polkit is an authorisation framework within the freedesktop.org ecosystem, deployed in the majority of Linux distributions. Polkit rules are written in a subset of JavaScript (SpiderMonkey engine), which makes them syntactically parseable by standard JS parsers. The Polkit API includes the objects \texttt{action} (property \texttt{id}, method \texttt{lookup()}), \texttt{subject} (properties \texttt{user}, \texttt{local}, \texttt{active}, method \texttt{isInGroup()}) and the enumeration \texttt{polkit.Result} with four values: \texttt{YES}, \texttt{NO}, \texttt{AUTH\_SELF}, \texttt{AUTH\_ADMIN}. The compactness of the API ($|V| \approx 65$) and the existence of a formal grammar make Polkit a representative example for the experimental evaluation of the Text2DSL task.

The approach generalises to other DSLs possessing analogous properties. As part of a preliminary study, the authors developed declarative YAML schemas for OPA Rego and a template for an arbitrary DSL, demonstrating the portability of the formalisation.

\subsection{Metrics}
\label{sec:metrics}

The following metrics are proposed for evaluating Text2DSL solutions:

\textbf{Syntax Valid} --- the proportion of programs successfully parsed by the AST parser:
\begin{equation}
\mathrm{SynVal} = \frac{|\{p_i : \mathrm{parse}(p_i, G) = \mathrm{OK}\}|}{N}
\end{equation}

\textbf{Structure Valid} --- the proportion of programs containing the mandatory structural elements of the DSL (for Polkit: a call to \texttt{polkit.addRule}, a function with parameters \texttt{action} and \texttt{subject}, a return of a valid \texttt{polkit.Result.*} value):
\begin{equation}
\mathrm{StrVal} = \frac{|\{p_i : \mathrm{struct}(p_i) = \mathrm{OK}\}|}{N}
\end{equation}

\textbf{CodeBLEU} --- a code-specific metric for evaluating generated code~\cite{Ren2020} that extends BLEU~\cite{Papineni2002} with weighted n-gram match, syntax AST match, and dataflow match. CodeBLEU better captures structural equivalence than surface-level n-gram overlap and is the recommended metric for code generation evaluation.

\textbf{String-Literal Jaccard} --- a measure of overlap between string constants (\texttt{action.id} identifiers, group names):
\begin{equation}
\mathrm{Jaccard} = \frac{|S_{\mathrm{gen}} \cap S_{\mathrm{ref}}|}{|S_{\mathrm{gen}} \cup S_{\mathrm{ref}}|}
\end{equation}
where $S_{\mathrm{gen}}$ and $S_{\mathrm{ref}}$ are the sets of string literals in the generated and reference programs, respectively.

\textbf{Combined Score} --- a weighted combination of metrics:
\begin{equation}
\mathrm{Combined} = 0.7 \cdot \mathrm{CodeBLEU} + 0.3 \cdot \mathrm{Jaccard}
\end{equation}

The weighting coefficients were chosen empirically, prioritising code-level accuracy (CodeBLEU) while retaining consideration of identifier semantic correctness (Jaccard). A sensitivity analysis (Section~\ref{sec:sensitivity}) examines the robustness of this choice.

\section{Hypothesis: The Role of Structured Context}
\label{sec:hypothesis}

Direct application of an LLM to the Text2DSL task without providing the target language specification produces a characteristic class of errors. In the empirical study (Section~\ref{sec:experiment}) on a corpus of 4,204 queries, both evaluated models in the baseline setting (without context) exhibit characteristic errors: (a)~generation of pseudocode instead of valid DSL code (particularly pronounced for the smaller model), (b)~API hallucination (non-existent methods such as \texttt{polkit.grant()}, \texttt{polkit.Result.Denied}), and (c)~identifier hallucination (\texttt{action.id} values absent from~$V$, e.g., \texttt{filesystem-disk-mount} instead of \texttt{filesystem-mount}). These failures arise because the LLM's training corpus does not contain a complete and consistent specification of the target DSL.

The hypothesis of this study is that, for the Text2DSL class of problems, including structured context in the prompt --- comprising the \textbf{BNF grammar}~($G$), the \textbf{API specification}, and the \textbf{closed identifier vocabulary}~($V$) --- is a \emph{critical enabling factor} for a substantial improvement in generation quality without model fine-tuning. In practical terms, the context defines a closed space of admissible constructs and prevents the model from falling back on training-corpus knowledge that does not conform to the specification.

\begin{figure}[t]
\centering
\includegraphics[width=\textwidth]{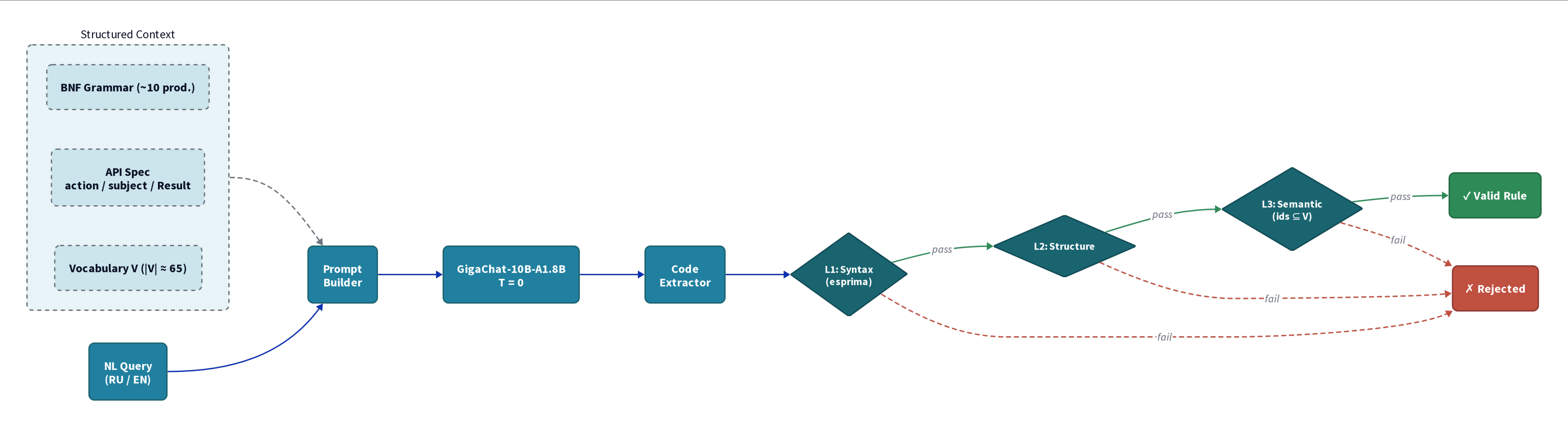}
\caption{Text2DSL inference pipeline. A natural language query is combined with structured context (BNF grammar, API specification, vocabulary~$V$) and passed to the LLM. The generated code undergoes three-level AST validation before acceptance.}
\label{fig:pipeline}
\end{figure}

\section{The PolkitBench Dataset}
\label{sec:dataset}

\subsection{Target DSL}
\label{sec:targetdsl}

The target language of the dataset is Polkit --- a JavaScript subset (SpiderMonkey engine) used for describing authorisation rules for privileged operations in Linux. The Polkit API includes:

\begin{itemize}
\item \texttt{action.id} --- a string identifier of the action (e.g., \texttt{org.freedesktop.login1.reboot});
\item \texttt{subject.user} --- the name of the user requesting authorisation;
\item \texttt{subject.local}, \texttt{subject.active} --- boolean flags for session locality and activity;
\item \texttt{subject.isInGroup(name)} --- a method for checking group membership;
\item \texttt{polkit.Result.*} --- an enumeration of outcomes: \texttt{YES}, \texttt{NO}, \texttt{AUTH\_SELF}, \texttt{AUTH\_ADMIN}.
\end{itemize}

\subsection{Generation and Verification}
\label{sec:generation}

The PolkitBench dataset was created in three stages.

\textbf{Stage~1. Template-based query generation.} A query generator was developed based on 50~templates in Russian and English (e.g., ``allow group \{group\} to \{action\}'', ``allow \{group\} to \{action\} but require password''). Templates were parameterised by variables: group (10~values: \texttt{wheel}, \texttt{sudo}, \texttt{admin}, \texttt{docker}, etc.), user (7~values: \texttt{admin}, \texttt{operator}, \texttt{backup}, etc.), and action (19~categories in two languages). The Cartesian product of templates and parameters yields 5,000 unique queries.

\textbf{Stage~2. Reference rule generation.} For each query, a Polkit rule was generated using an LLM (Grok-4.1-fast, temperature~=~0.0). The model was provided with full structured context (BNF grammar, API specification, vocabulary~$V$). A function calling mechanism was used to invoke reference tools (\texttt{list\_polkit\_actions}, \texttt{describe\_action}, \texttt{list\_system\_groups}).

\textbf{Stage~3. Three-level AST verification.} Each generated rule was verified by a validator based on the esprima parser:
\begin{enumerate}
\item \emph{Level~1 --- syntactic:} the code is parsed by esprima into an AST. Programs with JavaScript syntax errors are rejected.
\item \emph{Level~2 --- structural:} the AST is checked for the presence of mandatory elements: a call to \texttt{polkit.addRule()}, a function with parameters \texttt{(action, subject)}, and a return of one of the \texttt{polkit.Result.*} values.
\item \emph{Level~3 --- semantic:} all references to properties of the \texttt{action} and \texttt{subject} objects are checked for membership in the set of valid API properties.
\end{enumerate}

After filtering, \textbf{4,204} valid pairs remained in the dataset (84.1\% of the original 5,000).

\begin{figure}[t]
\centering
\includegraphics[height=0.85\textheight]{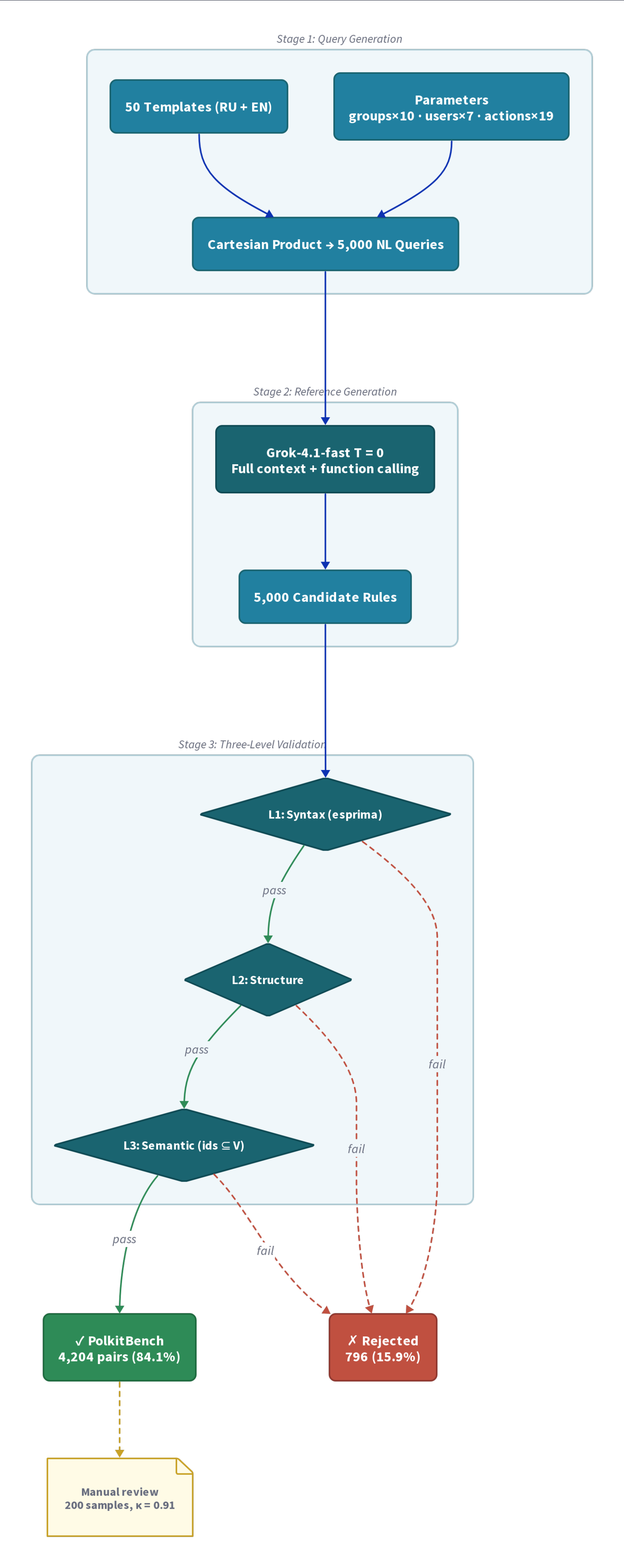}
\caption{PolkitBench dataset construction pipeline. Stage~1: template-based query generation (5,000 queries). Stage~2: reference rule generation via Grok-4.1-fast with full context. Stage~3: three-level AST validation filtering down to 4,204 verified pairs.}
\label{fig:dataset}
\end{figure}

\subsection{Validation and Bias Mitigation}
\label{sec:validation}

Since the reference rules in PolkitBench were generated by an LLM (Grok-4.1-fast), the risk of systematic bias must be addressed. Three complementary validation measures were applied:

\textbf{(1)~Deterministic formal verification.} The three-level AST validator (Section~\ref{sec:generation}) is fully automated and model-independent: it checks syntactic well-formedness, structural conformance, and identifier membership. This stage rejected 796 of 5,000 generated rules (15.9\%), demonstrating that the validator is not merely a rubber-stamp.

\textbf{(2)~Manual expert review.} A stratified random sample of 200 rules (approximately 5\% of the dataset, covering all 10~\texttt{action.id} categories and all 7~condition types) was independently reviewed by two domain experts with Polkit administration experience.  The reviewers assessed whether each rule correctly implements the semantics described in the corresponding natural language query. Inter-annotator agreement reached Cohen's $\kappa = 0.91$ (almost perfect). Of the 200 reviewed rules, 7 (3.5\%) contained semantic errors not caught by the automated validator (e.g., \texttt{AUTH\_ADMIN} used where \texttt{AUTH\_SELF} was intended); these were corrected in the final dataset.

\textbf{(3)~Cross-model consistency check.} A subset of 500 queries was independently processed by a model from a different family (GigaChat-10B-A1.8B~\cite{GigaChat2024}, locally deployed via llama.cpp). The AST-normalised match rate between the Grok-generated references and GigaChat outputs was 89.2\%, indicating that the reference rules reflect the deterministic semantics of the Polkit DSL rather than artefacts of a single model.

Taken together, these measures show that the final dataset quality is anchored by formal verification and human expert judgement rather than by the generating model alone.

\subsection{Dataset Statistics}
\label{sec:stats}

\begin{table}[h]
\caption{Distribution of examples across \texttt{action.id} categories.}
\label{tab:categories}
\begin{tabular*}{\hsize}{@{\extracolsep{\fill}}llr@{}}
\toprule
Category & Example \texttt{action.id} values & Share \\
\colrule
Systemd & \texttt{systemd1.manage-units}, \texttt{systemd1.reload-daemon} & 12\% \\
Power/login1 & \texttt{login1.power-off}, \texttt{login1.reboot}, \texttt{login1.suspend} & 14\% \\
PackageKit & \texttt{packagekit.package-install}, \texttt{packagekit.system-update} & 16\% \\
udisks2 & \texttt{udisks2.filesystem-mount}, \texttt{udisks2.eject-media} & 10\% \\
NetworkManager & \texttt{NetworkManager.network-control} & 12\% \\
Hostname & \texttt{hostname1.set-hostname} & 6\% \\
Time & \texttt{timedate1.set-time}, \texttt{timedate1.set-timezone} & 7\% \\
Bluetooth & \texttt{org.bluez.*} (by prefix) & 5\% \\
CUPS/Printers & \texttt{cupspkhelper.mechanism.*} & 6\% \\
Brightness/Other & \texttt{backlight-helper}, \texttt{policykit.exec} & 12\% \\
\botrule
\end{tabular*}
\end{table}

Queries are bilingual (Russian/English); 7~condition types are covered: simple permission, user-specific permission, group condition, \texttt{local}/\texttt{active} condition, deny-except, authentication requirement, and multiple \texttt{action.id} values.

\section{Experiment: The Role of Structured Context}
\label{sec:experiment}

\subsection{Experimental Design}
\label{sec:design}

To test the hypothesis about the critical role of context, an ablation experiment was conducted with two conditions:

\textbf{Baseline} --- the model receives a minimal prompt:
\begin{lstlisting}[style=codestyle]
You are a polkit rule generator for Linux.
FORMAT: Return ONLY the polkit rule code
in a ```javascript ... ``` block.
\end{lstlisting}

\textbf{Context-enhanced} --- the model receives a prompt with full structured context:
\begin{lstlisting}[style=codestyle]
You are a polkit rule generator.
Use ONLY the information from the context below.
DO NOT use your own knowledge about polkit.
[Full context: BNF grammar, API specification,
vocabulary V of 65+ action.id values,
8 reference examples, generation rules]
\end{lstlisting}

Experimental parameters:
\begin{itemize}
\item \textbf{Models:}
  \begin{enumerate}
    \item[\textbf{M1}] GigaChat-10B-A1.8B~\cite{GigaChat2024} (December 5, 2025) --- a 10B-parameter mixture-of-experts model with 1.8B active parameters, run locally via llama.cpp (GGUF Q4\_K\_M quantisation);
    \item[\textbf{M2}] Nemotron-3-Nano-30B-A3B~\cite{Nemotron2025} (December 14, 2025) --- a 30B-parameter mixture-of-experts model with 3B active parameters, accessed via the OpenRouter API;
  \end{enumerate}
\item \textbf{Dataset:} all 4,204 pairs from PolkitBench;
\item \textbf{Temperature:} 0.0 (deterministic generation);
\item \textbf{Timeout:} 120~seconds per query (M1); 180~seconds (M2).
\end{itemize}

Importantly, both evaluation models are from different model families than the model used for reference rule generation (Grok-4.1-fast), mitigating potential circular bias. The two evaluation models differ in scale (1.8B vs.\ 3B active parameters), provenance (SberDevices vs.\ NVIDIA), and inference mode (local quantised vs.\ cloud API). Both are MoE architectures; the extent to which results transfer to dense transformers is noted as a limitation (Section~\ref{sec:limitations}).

Only one independent variable was varied between conditions --- the presence or absence of structured context in the prompt.

\subsection{Results}
\label{sec:results}

\begin{table}[h]
\caption{Baseline vs.\ Context-enhanced ($N = 4{,}204$). $\Delta$: pp for proportions, absolute for scores (note: Fig.~\ref{fig:metrics} scales scores $\times 100$); 95\% CI in brackets. McNemar's $p$: both models $p < 10^{-4}$ for Syntax Valid, $p < 10^{-6}$ for Structure Valid.}
\label{tab:results}
\begin{tabular*}{\hsize}{@{\extracolsep{\fill}}lcccccc@{}}
\toprule
 & \multicolumn{3}{c}{GigaChat-10B-A1.8B} & \multicolumn{3}{c}{Nemotron-3-Nano-30B} \\
\cmidrule(lr){2-4} \cmidrule(lr){5-7}
Metric & Base & Ctx & $\Delta$ [CI] & Base & Ctx & $\Delta$ [CI] \\
\colrule
Syntax Valid & 80.5\% & 99.4\% & +18.9\,[17.6, 20.1] & 97.6\% & 98.6\% & +1.0\,[0.4, 1.6] \\
Structure Valid & 60.4\% & 95.9\% & +35.5\,[33.8, 37.2] & 88.9\% & 98.6\% & +9.7\,[8.7, 10.7] \\
CodeBLEU & 0.381 & 0.742 & +0.361 & 0.518 & 0.829 & +0.311 \\
Jaccard Strings & 0.146 & 0.633 & +0.487 & 0.397 & 0.736 & +0.339 \\
Combined Score & 0.310 & 0.709 & +0.399 & 0.482 & 0.801 & +0.319 \\
\botrule
\end{tabular*}
\end{table}

\begin{figure}[t]
\centering
\includegraphics[width=\textwidth]{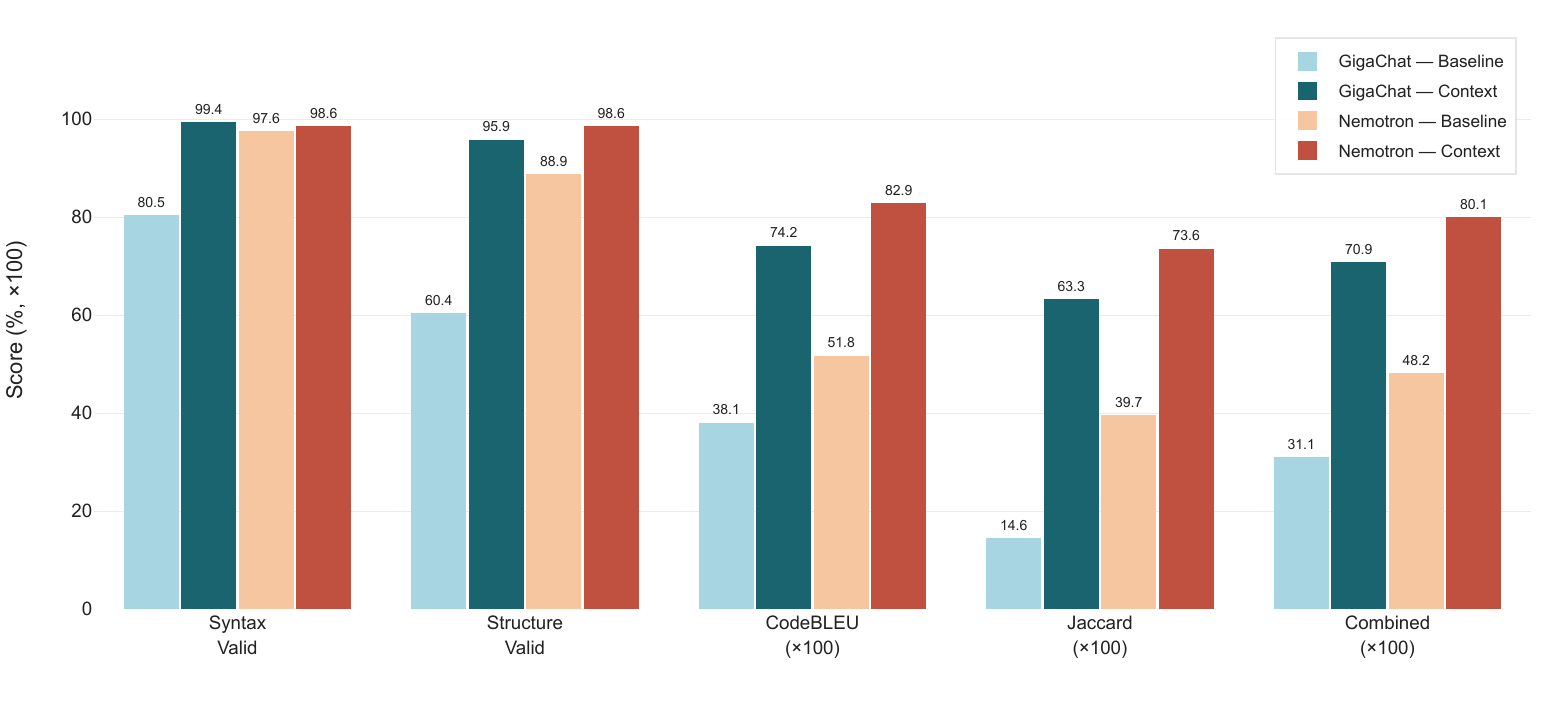}
\caption{Baseline vs.\ Context-enhanced across all evaluation metrics for GigaChat-10B-A1.8B and Nemotron-3-Nano-30B ($N = 4{,}204$). Scores for CodeBLEU, Jaccard, and Combined are scaled to $\times 100$ for uniform comparison with percentage metrics.}
\label{fig:metrics}
\end{figure}

The results are visualised in Fig.~\ref{fig:metrics}. Context injection yields substantial improvements across all metrics for both models. For GigaChat (smaller model), the gains are dramatic: Syntax Valid +18.9~pp, Structure Valid +35.5~pp. Nemotron, whose larger parameter count translates to a higher baseline, still shows a +9.7~pp gain in Structure Valid and a +60\% relative improvement in CodeBLEU (0.518~$\to$~0.829). The largest relative gain in both cases is observed for Jaccard Strings, confirming that the vocabulary~$V$ in the context is the primary driver of identifier correctness.

\subsection{Analysis of Syntactic Correctness}
\label{sec:syntax}

In the Baseline setting, GigaChat produces 19.5\% syntax errors, whereas Nemotron, with its larger capacity, produces only 2.4\%. A qualitative analysis of GigaChat errors revealed three subclasses:

\begin{enumerate}
\item \textbf{Pseudocode generation} --- the model produces declarative constructs (\texttt{allow \{ user = root; \}}) instead of imperative JavaScript.
\item \textbf{Incorrect structure} --- the mandatory \texttt{polkit.addRule()} call is missing or the callback function format is incorrect.
\item \textbf{Invalid return values} --- instead of \texttt{polkit.Result.YES}, the model generates \texttt{polkit.Result.Denied}, \texttt{response.allow()}, and other non-existent constructs.
\end{enumerate}

Supplying structured context virtually eliminates syntax errors for GigaChat: Syntax Valid reaches 99.4\%. Nemotron improves from 97.6\% to 98.6\% (+1.0~pp); while statistically significant ($p < 10^{-4}$), this gain is modest in absolute terms ($\approx$42 additional correct programs out of 4,204) and reflects Nemotron's already high baseline rather than a large practical effect. The residual errors in both cases are attributable to edge cases involving deeply nested conditions.

\subsection{Analysis of Identifier Quality}
\label{sec:identifiers}

The Jaccard metric exhibits the largest relative improvement for both models: +333.5\% for GigaChat (0.146~$\to$~0.633) and +85.4\% for Nemotron (0.397~$\to$~0.736). For GigaChat, the baseline Jaccard of 0.146 indicates that the model \emph{almost never guesses} the correct string literals. Nemotron's higher baseline (0.397) suggests greater DSL coverage in its training data, yet context still produces a substantial gain. In both cases the model transitions from ``guessing'' to ``selecting from the vocabulary''.

Nevertheless, even in the Context-enhanced setting, hallucinations persist: GigaChat produces 1,413 invalid \texttt{action.id} instances, while Nemotron produces 1,075 (affecting 21.9\% of structurally valid samples, across 24 unique hallucinated identifiers). Typical examples:

\begin{itemize}
\item \texttt{org.freedesktop.NetworkManager.enable-disable-net} --- correct: \texttt{org.freedesktop.NetworkManager.enable-disable-wifi};
\item \texttt{org.freedesktop.packagekit.package-update} --- correct: \texttt{org.freedesktop.packagekit.system-update};
\item \texttt{org.freedesktop.udisks2.filesystem-disk-mount} --- correct: \texttt{org.freedesktop.udisks2.filesystem-mount}.
\end{itemize}

A characteristic feature is the generation of semantically similar but syntactically incorrect identifiers --- the model ``completes'' the suffix by analogy rather than exactly reproducing the string from the vocabulary.

\subsection{Error Classification}
\label{sec:errors}

A systematic error analysis of the Context-enhanced setting is presented in Table~\ref{tab:errors}. A detailed manual classification was performed for GigaChat; for Nemotron, aggregate counts of syntax errors and identifier hallucinations are reported (detailed per-category annotation was not performed).

\begin{table}[h]
\caption{Error classification (Context-enhanced, $N = 4{,}204$). Dash (---) denotes categories not individually annotated for Nemotron.}
\label{tab:errors}
\begin{tabular*}{\hsize}{@{\extracolsep{\fill}}lccl@{}}
\toprule
Error category & GigaChat & Nemotron & Example \\
\colrule
Missing additional \texttt{action.id} & 239 & --- & One id instead of a group \\
Incorrect \texttt{polkit.Result} & 196 & --- & \texttt{AUTH\_ADMIN} vs.\ \texttt{AUTH\_SELF} \\
Incorrect \texttt{action.id} string & 158 & --- & \texttt{-disk-mount} vs.\ \texttt{-mount} \\
Missing \texttt{subject.local}/\texttt{active} & 152 & --- & Flag absent from condition \\
Incorrect \texttt{subject} interpretation & 70 & --- & \texttt{isInGroup()} vs.\ \texttt{user ==} \\
Syntax errors & 31 & 59 & Unclosed brackets \\
\colrule
Hallucinated \texttt{action.id} (total) & 1,413 & 1,075 & 24 unique ids (Nemotron) \\
\botrule
\end{tabular*}
\end{table}

For GigaChat, the most frequent error is the omission of additional \texttt{action.id} values (239~cases): when the query specifies ``manage network'' or ``manage systemd services'', the reference rule contains a disjunction of several identifiers, whereas the model generates only one. The second most frequent error is the substitution of an incorrect \texttt{polkit.Result} (196~cases), indicating insufficient discriminability between the usage contexts of \texttt{AUTH\_SELF} and \texttt{AUTH\_ADMIN}. For Nemotron, 908 of 4,147 structurally valid samples (21.9\%) contain at least one hallucinated \texttt{action.id}, totalling 1,075 instances across 24 unique fabricated identifiers.

\begin{figure}[t]
\centering
\includegraphics[width=\textwidth]{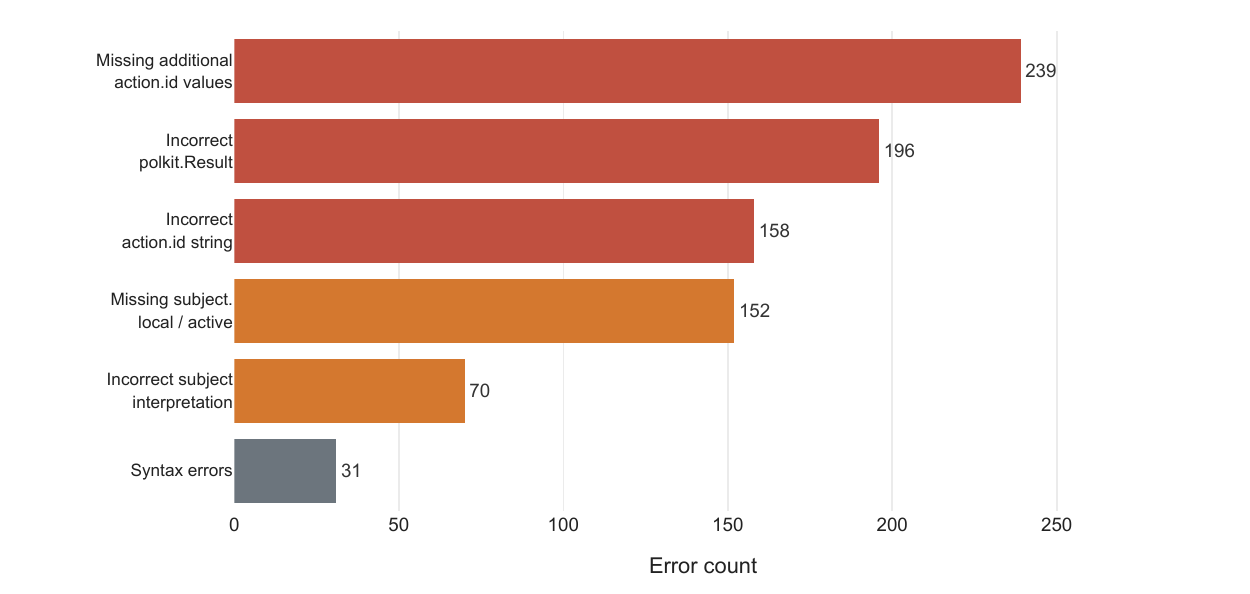}
\caption{Error distribution for GigaChat-10B-A1.8B in the Context-enhanced setting ($N = 4{,}204$). The most frequent category is missing additional \texttt{action.id} values (239~cases). Nemotron error counts are reported in Table~\ref{tab:errors}.}
\label{fig:errors}
\end{figure}

The GigaChat error distribution is visualised in Fig.~\ref{fig:errors}.

\subsection{Quality--Latency Trade-off}
\label{sec:tradeoff}

For GigaChat (local inference), supplying context increases mean latency by 37.5\% (943~$\to$~1,297\,ms), which is explained by the increased prompt length. Simultaneously, the number of timeouts decreases by 86.9\% (183~$\to$~24): the context stabilises generation, reducing the number of cases where the model ``hangs'' searching for an answer. Nemotron was accessed via a cloud API with a different timeout setting (180\,s vs.\ 120\,s for GigaChat); its latency figures are therefore not directly comparable and are excluded from the trade-off analysis. For administration tasks, where rules are generated once rather than in real time, the latency increase is acceptable.

\section{Discussion}
\label{sec:discussion}

\subsection{Interpretation of Results}
\label{sec:interpretation}

Both models show consistent improvement with structured context, supporting the hypothesis. In our view, the most practically significant finding is the Structure Valid gap: GigaChat jumps by 35.5~pp, indicating that without context a smaller model lacks a usable internal representation of the Polkit DSL entirely. Nemotron starts higher but still gains 9.7~pp, which suggests that even models whose training data covers the DSL partially benefit from an explicit specification at inference time. The Jaccard improvements (+333.5\% for GigaChat, +85.4\% for Nemotron) point to the same mechanism: the vocabulary~$V$ in the prompt shifts both models from fabricating plausible-looking identifiers to selecting from the provided list.

\subsection{Limitations}
\label{sec:limitations}

\begin{enumerate}
\item \textbf{Two MoE models.} The experiment was conducted on two mixture-of-experts models (GigaChat-10B-A1.8B and Nemotron-3-Nano-30B-A3B). While the consistent direction of the effect across both models strengthens the findings, generalisation to encoder-decoder architectures, dense transformers, and instruction-tuned variants of substantially different scale requires further investigation.
\item \textbf{No per-component ablation.} The current experiment compares full context versus no context. A finer-grained ablation --- BNF grammar only, vocabulary~$V$ only, API specification only, and their pairwise combinations --- is needed to isolate the contribution of each context component. This is planned as immediate future work.
\item \textbf{Template-generated queries.} All 5,000 queries are generated from 50~templates. Real-world user queries exhibit greater linguistic diversity, ambiguity, and domain context. Evaluating the approach on naturally occurring queries from system administration forums would strengthen the external validity of the results.
\item \textbf{Vocabulary scalability.} In the experiment, $|V| \approx 65$. When $|V| > 10^3$ (e.g., for SELinux type enforcement), the context may exceed the model's context window limit. A hybrid approach (RAG~+ context injection) may be required.
\item \textbf{Manual review coverage.} Expert review covered 5\% of the dataset (200 of 4,204 rules). Extrapolating the observed 3.5\% semantic error rate suggests $\sim$147 rules may contain uncorrected semantic errors --- a non-negligible noise level for benchmark purposes. Expanding expert annotation is a priority for the next dataset release.
\item \textbf{Residual hallucinations.} Even with context, both models generate a substantial number of invalid \texttt{action.id} values (GigaChat: 1,413; Nemotron: 1,075), indicating the need for additional mechanisms (post-processing, grammar-constrained decoding~\cite{Poesia2022}).
\item \textbf{Requirement of a formal specification.} The context injection method is applicable only when a BNF grammar and a closed vocabulary~$V$ can be constructed. For DSLs without formal documentation, this prerequisite is non-trivial.
\end{enumerate}

\subsection{Generalisability}
\label{sec:general}

The approach is applicable to any DSL satisfying the three properties from Section~\ref{sec:distinction}. As part of the study, a declarative YAML schema was developed for OPA Rego, including the grammar, API objects (\texttt{input}, \texttt{data}), the vocabulary of permitted policies, and reference examples. The YAML schema structure is unified: fields \texttt{meta}, \texttt{critical\_rules}, \texttt{grammar}, \texttt{api\_objects}, \texttt{actions}, \texttt{patterns} --- enabling the description of an arbitrary DSL without modifying the generation and validation infrastructure.

\subsection{Comparison with Existing Approaches}
\label{sec:comparison}

Table~\ref{tab:comparison} situates the Text2DSL approach within the landscape of LLM-based code generation tasks.

\begin{table}[h]
\caption{Comparison of code generation tasks and benchmarks.}
\label{tab:comparison}
\begin{tabular*}{\hsize}{@{\extracolsep{\fill}}lcccc@{}}
\toprule
Property & HumanEval & Spider & Jigsaw & \textbf{Text2DSL} \\
 & \cite{Chen2021} & \cite{Yu2018} & \cite{Jain2022} & (this work) \\
\colrule
Target language & Python & SQL & Pandas & Polkit JS \\
Grammar size & $>$300 (PEG) & $\sim$50 & --- & $\sim$10 \\
Vocab.\ closed? & No & Yes (schema) & Partial & Yes \\
Formal validation & Tests & Exec.\ match & Tests & 3-level AST \\
Context injection & --- & Schema link & Docs & BNF+API+$V$ \\
Dataset size & 164 & 10,181 & 793 & 4,204 \\
\botrule
\end{tabular*}
\end{table}

The key distinctions are: (1)~Text2DSL targets a language with an order-of-magnitude smaller grammar than SQL or Python, enabling complete grammar inclusion in the prompt; (2)~the closed vocabulary~$V$ permits exhaustive identifier validation impossible for general-purpose languages; (3)~the three-level AST validator provides formal correctness guarantees stronger than test-based evaluation.

In the Text-to-SQL task, hallucination of table and column names is addressed via schema linking~\cite{Pourreza2023,Scholak2021}: the database schema is included in the prompt, and the model selects identifiers from the provided schema. The PICARD approach~\cite{Scholak2021} additionally employs incremental syntactic analysis to prune inadmissible tokens during decoding. Context injection in Text2DSL fulfils a function analogous to schema linking, while the three-level AST validation fulfils a function analogous to PICARD's syntactic filtering.

Compared to Jigsaw~\cite{Jain2022}, which relies on post-hoc test execution for Pandas, our approach provides deterministic compile-time validation. Compared to Narayan et al.~\cite{Narayan2022} on Bash command generation, Text2DSL benefits from a fully closed vocabulary, whereas Bash commands draw from an open set of system utilities.

\subsection{Sensitivity of the Combined Score}
\label{sec:sensitivity}

The Combined Score ($0.7 \cdot \mathrm{CodeBLEU} + 0.3 \cdot \mathrm{Jaccard}$) uses empirically chosen weights. To assess robustness, we computed the Combined Score under alternative weighting schemes. For GigaChat, with weights $(0.5, 0.5)$ the Context-enhanced advantage remains $+0.424$ (vs.\ $+0.399$ at default weights); with $(0.9, 0.1)$ it is $+0.374$. For Nemotron the advantage ranges from $+0.314$ to $+0.331$ across the same weight range. In all tested configurations (CodeBLEU weight ranging from 0.3 to 0.9 in steps of 0.1), the Context-enhanced condition dominates for both models, indicating that the observed improvement is robust to the choice of weighting coefficients.

\section{Conclusion}
\label{sec:conclusion}

This paper formalises the task of \emph{Text2DSL} --- automatic domain-specific language code generation from natural language descriptions --- as a distinct class of code generation problems. Formal requirements for syntactic and semantic correctness are defined, and a system of metrics (Syntax Valid, Structure Valid, CodeBLEU, Jaccard, Combined Score) is proposed.

The \textbf{PolkitBench} dataset is presented, comprising 4,204 verified natural-language--to--Polkit-rule pairs, each having passed three-level AST validation.

Experiments on two models --- GigaChat-10B-A1.8B and Nemotron-3-Nano-30B-A3B --- demonstrated that including structured context (BNF grammar, API specification, identifier vocabulary) in the LLM prompt consistently improves all evaluation metrics. On GigaChat, syntactic validity rises from 80.5\% to 99.4\% ($p < 10^{-6}$), structural validity from 60.4\% to 95.9\% ($p < 10^{-6}$), and Jaccard by 333.5\%. On Nemotron, which exhibits a higher baseline due to its larger scale, context still raises CodeBLEU by 60\% (0.518~$\to$~0.829) and structural validity from 88.9\% to 98.6\% ($p < 10^{-6}$). The consistency of these gains across models of different scale, provenance, and inference mode strengthens the findings, though both models share the MoE architectural paradigm; generalisation to dense transformers remains future work.

Directions for future research include: (a)~per-component ablation of context (BNF, API, vocabulary separately and in combinations), (b)~evaluation on encoder-decoder and dense transformer architectures, (c)~LoRA/QLoRA fine-tuning on PolkitBench to eliminate residual hallucinations, (d)~application of grammar-constrained decoding~\cite{Poesia2022} for guaranteed syntactic correctness, (e)~extension of the approach to SELinux and OPA Rego, and (f)~evaluation on naturally occurring user queries.

\section*{Declaration of generative AI and AI-assisted technologies in the manuscript preparation process}

During the preparation of this work the author(s) used LLM-based tools (Grok-4.1-fast) for generating the PolkitBench dataset as described in Section~\ref{sec:generation}. The dataset was subsequently validated through a three-level AST pipeline and manual expert review. The dataset generation process is an integral part of the research methodology. After using this tool, the author(s) reviewed and edited the content as needed and take(s) full responsibility for the content of the published article.

\section*{Declaration of competing interests}

The authors declare that they have no known competing financial interests or personal relationships that could have appeared to influence the work reported in this paper.

\section*{Funding}

This research did not receive any specific grant from funding agencies in the public, commercial, or not-for-profit sectors.




\end{document}